\documentclass[a4paper, 10pt, conference]{ieeeconf}      
\usepackage{FG2021}
\usepackage{graphicx}
\usepackage{array, makecell} %
\usepackage{tabularx,booktabs}
\usepackage{subcaption}

\FGfinalcopy 

\IEEEoverridecommandlockouts                              
\def\FGPaperID{9} 

\title{\LARGE \bf
FocusFace: Multi-task Contrastive Learning for Masked Face Recognition
}

\author{\parbox{16cm}{\centering
    {\large Pedro C. Neto$^{1,2}$, Fadi Boutros$^{3,4}$, João Ribeiro Pinto$^{1, 2}$,\\ Naser Damer$^{3,4}$ , Ana F. Sequeira $^{1}$ and Jaime~S.~Cardoso$^{2, 1}$}\\
    {\normalsize
    $^1$ INESC TEC, Porto, Portugal\\
    $^2$ Faculdade de Engenharia da Universidade do Porto, Porto, Portugal\\
    $^3$ Fraunhofer Institute for Computer Graphics Research IGD, Darmstadt, Germany\\
    $^4$ Department of Computer Science, TU Darmstadt, Darmstadt, Germany}}
    \thanks{This work was financed by National Funds through the Portuguese funding agency, FCT - Fundação para a Ciência e a Tecnologia within project UIDB/50014/2020, and within the PhD grants ``2021.06872.BD'' and ``SFRH/BD/137720/2018''. This research work has been also funded by the German Federal Ministry of Education and Research and the Hessen State Ministry for Higher Education, Research and the Arts within their joint support of the National Research Center for Applied Cybersecurity ATHENE.}
}

\begin{document}

\ifFGfinal
\thispagestyle{empty}
\pagestyle{empty}
\else
\author{Anonymous FG4COVID19 2021 submission\\ Paper ID \FGPaperID \\}
\pagestyle{plain}
\fi
\maketitle

\begin{abstract}

SARS-CoV-2 has presented direct and indirect challenges to the scientific community. One of the most prominent indirect challenges advents from the mandatory use of face masks in a large number of countries. Face recognition methods struggle to perform identity verification with similar accuracy on masked and unmasked individuals. It has been shown that the performance of these methods drops considerably in the presence of face masks, especially if the reference image is unmasked. We propose FocusFace, a multi-task architecture that uses contrastive learning  to be able to accurately perform masked face recognition. The proposed architecture is designed to be trained from scratch or to work on top of state-of-the-art face recognition methods without sacrificing the capabilities of a existing models in conventional face recognition tasks. We also explore different approaches to design the contrastive learning module. Results are presented in terms of masked-masked (M-M) and unmasked-masked (U-M) face verification performance. For both settings, the results are on par with published methods, but for M-M specifically, the proposed method was able to outperform all the solutions that it was compared to. We further show that when using our method on top of already existing methods the training computational costs decrease significantly while retaining similar performances. The implementation and the trained models are available at GitHub. 

\end{abstract}

\section{Introduction}

The current SARS-CoV-2 pandemic and related hygienic concerns intensified the necessity to adopt contactless biometric recognition systems. One of the most common contactless biometric recognition systems that meet high accuracy requirements is face recognition. It has been widely adopted in several scenarios such as automated border control, access control of private areas and convenience applications~\cite{gorodnichy2014automated,lovisotto2017mobile}. Nonetheless, the pandemic was also responsible for drastically increasing the usage of face masks. These masks, enforced by several governments, are a form of face occlusion and limit the performance of face recognition systems. The most recent studies on the effects of the face mask on face recognition accuracy have indicated a significant degradation in the accuracy of these models~\cite{DamerBiosig2020,ngan2020ongoinga,ngan2020ongoingb,JEEVAN2022108308}. 

The new scenario imposed by the pandemic situation motivated the development of face recognition approaches robust to face masks. We propose a multi-task learning constraint approach to extend ArcFace loss~\cite{deng2019arcface}. Current works focus on the addition of synthetic masks to the original dataset~\cite{anwar2020masked} or restraining the input to contain only the periocular area of the face~\cite{li2021cropping}. Our approach promotes the focus on uncovered areas of the face. By combining cross-entropy and ArcFace loss into a multi-task loss for joint mask detection and face recognition we promote a behavior similar to the one shown by periocular methods. However, we do not restrain the available information to be exclusively from the periocular area. Hence, we argue that our method learns where to focus in a end-to-end fashion. The proposed method is trained with both masked and unmasked versions of an image. For each of those images, the network predicts the corresponding identity and whether the person is wearing a mask. The focus on unmasked areas is promoted by minimizing the distance between the masked and unmasked embeddings produced by the network in a contrastive learning fashion. Moreover, as mentioned by Boutros~\textit{et al.}~\cite{boutros2021mfr}, the vast majority of current methods are focused on training models from scratch. However, ours can also be utilized on top of state-of-the-art face recognition methods.  

The proposed method is trained and validated on a dataset using synthetic masks~\cite{deng2019arcface,huang2008labeled}, and evaluated on a dataset that contains real masks~\cite{DamerBiosig2020,DamerIetbmt2021}. The latter was also used to evaluate the solutions of the International Joint Conference on Biometrics (IJCB 2021) competition on masked face recognition. The verification results support the performance gains of our approach with an FMR100 (the lowest false non-match rate (FNMR) for false match rate (FMR) $\le$ 1.0 \%) improvement over the regular ArcFace loss. 

This paper is further divided into four major sections. Section~\ref{sec:related_work} discusses related work focused on masked face recognition. Afterwards, our approach to tackling the problem is presented in Section~\ref{sec:methods}. The implementation details and experimental setup are described in Section~\ref{sec:experimental_setup}. Section~\ref{sec:results} compares the achieved results and discusses their meaning. Finally, the conclusions are drawn in Section~\ref{sec:conclusion}.


\section{Related Work}
\label{sec:related_work}

\begin{figure*}[!h]
\centering
\includegraphics[width=0.635\textwidth]{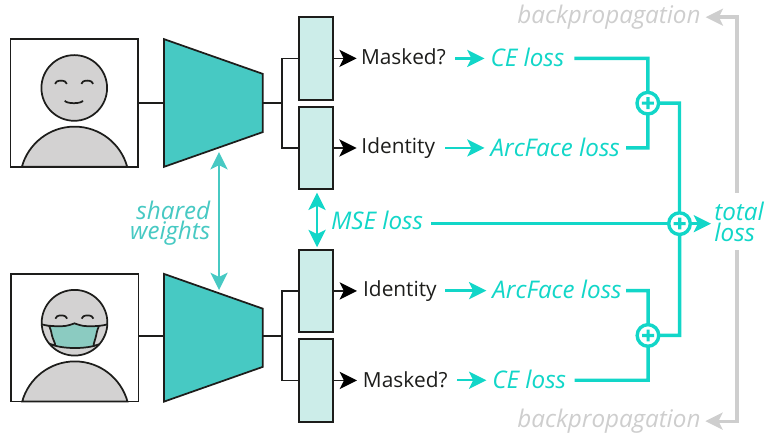}
\caption{Architecture of the proposed model. The backbone network receives masked and unmasked inputs, and computes the two embeddings for each input. The smaller embedding of each input is trained for mask detection with the cross-entropy (CE) loss. Whereas the other is trained for face recognition with the ArcFace loss. A third component of the loss is also included in the form of the mean squared error (MSE) between the two embeddings used for face recognition. }
\label{architecture_scheme}
\end{figure*}

Recent research on face recognition made significant improvements in the verification performance on non-occluded faces \cite{deng2019arcface,boutros2021elasticface,DBLP:conf/icb/BoutrosDFKK21}. Before the COVID-19 pandemic, the works on occluded faces were rather limited and scarce, and focused on sunglasses or scarves~\cite{opitz2016grid,song2019occlusion}. Following the wide adoption of face masks during the pandemic situation, several studies examined the effects of wearing a mask on face recognition \cite{DBLP:journals/corr/abs-2102-09258}. Two important studies were conducted by the National Institute of Standards and Technology (NIST). The first evaluated pre-COVID-19 algorithms on masked face recognition~\cite{ngan2020ongoinga}, whereas the second evaluated post-COVID-19 algorithms~\cite{ngan2020ongoingb}. These studies are a component of the ongoing Face Recognition Vendor Test (FRVT), and indicated that competitive algorithms on unmasked faces, which fail to authenticate less than 1\% of the individuals, have their failure rate increased up to 50\% in the presence of masks. Despite the conclusions of these studies indicating that face masks hurt the performance of face recognition, their validity is limited due to the exclusive use of synthetic masks. Hence, it might not capture the real effect of wearing masks. Damer~\textit{et al.}~\cite{DamerBiosig2020} have additionally studied the limitations caused by face masks on two academic face recognition algorithms and one commercial solution. In their study, the evaluation dataset was collected over three sessions with 24 participants. Both Damer~\textit{et al.}~\cite{DamerBiosig2020} and a Department of Homeland Security (DHS) study~\cite{JEEVAN2022108308} showed the performance degradation that occurs whenever face masks are worn. The first has shown that the SphereFace method can have its FMR100 increased from $0.065\%$ to $27.35\%$ when comparing conventional face recognition to masked face recognition. Wearing a mask have been additionally shown to significantly affect the verification performance of human operators \cite{DBLP:journals/corr/abs-2103-01924}, as well as other components of face recognition systems like face quality estimation \cite{FuMaskedQuality} and presentation attack detection \cite{FANG2021108398}.

The most active research area related to facial images and face masks is the detection of the latter. Several articles focused on this task have been published~\cite{loey2021hybrid,qin2020identifying}. Nevertheless, these studies present virtually no improvement to the task of masked face recognition. Neither focused on evaluating the impact of the mask on face recognition. The work presented by Li~\textit{et al.}~\cite{li2021cropping}, which proposed an attention-based method to train a model on the periocular area of masked faces, has shown interesting improvements. However, its evaluation was limited by the exclusive use of simulated masks. Other works with limited testing, either due to the use of simulated masks or small datasets, have been published~\cite{anwar2020masked}. The experiments focused mostly on fine-tuning pre-trained models, such as FaceNet~\cite{schroff2015facenet}, with simulated masks. 

In an attempt to mitigate these limitations, Boutros~\textit{et al.}~\cite{boutros2021mfr} have promoted a competition within the 2021 International Joint Conference on Biometrics (IJCB). The works that joined the competition were evaluated on a dataset containing real masks and a significant number of samples (1,511,113 masked-masked pairs and 800,207 unmasked-masked pairs)~\cite{DamerBiosig2020,DamerIetbmt2021}. The vast majority of the competing solutions relied on retraining the ArcFace model with datasets containing simulated masks. One of the approaches proposed multi-task learning to jointly detect masks and identify the correct sample identity~\cite{montero2021boosting}. That multi-task model, despite learning to capture masked related patterns, did not have a mechanism to promote the use of that information to improve face recognition by avoiding masked areas. Neto~\textit{et al.}~\cite{neto2021my} proposed a learning strategy based on a modified triplet loss that promoted the network focus on specific areas of the face. Besides the competition-related work, there have also been methodologies based on modifications to the triplet loss~\cite{boutros2021unmasking}.

Motivated by the endeavours of the solutions proposed to the IJCB 2021 competition and by the recent evaluations on the harmful effect of face masks on face recognition performance, we present in this work a novel approach to improve masked face recognition. Differently from previous approaches, our multi-task method is evaluated on a large test set and contains mechanisms to explore the learnt features related to the mask in order to improve masked face recognition. The proposed solution can be used on top of existing solutions or trained from scratch, thus it does not require the sacrifice of conventional face recognition performance. The details are presented in Section~\ref{sec:methods}.

\section{Methodology}
\label{sec:methods}

The proposed approach is composed of two distinct components designed to work together. First, the solution delves into a multi-task learning approach so that the network becomes aware of the existence of a face mask. Secondly, mask awareness promotes a more stable learning of a contrastive loss between the features of both masked and unmasked images. Our approach is further detailed in the following paragraphs of this section. 

\subsection{Multi-task architecture}

The proposed architecture contemplates two distinct tasks. First, the most important of both is the recognition and classification of the image identity. The other is responsible for the detection of a face mask in the input image.  The architecture uses a convolutional neural network to produce two distinct embeddings, one per task. These embeddings are then optimized for their respective tasks. 

By adding a component to the network to improve its capabilities to distinguish between masked and unmasked individuals, it is created a mechanism to promote mask awareness. This awareness is intended to improve the convergence stability and the features extracted. For this, the embedding dedicated to the detection of the mask is given to a fully connected layer with one output neuron. The optimization for this task is carried by minimizing the cross-entropy loss. The formulation of this loss is seen in Equation~\ref{eq:ce}, where $N$ represents the number of samples in the mini-batch, $n$ is the class number, and  $y_{it}$ represents the output of the network for the sample $i$ and target class $t$.

\begin{equation}
    L_{CE} = -\frac{1}{N}\sum_{i\in N}{\log\frac{e^{y_{it}}}{\sum_{k\in n}e^{y_{ik}}}}
\label{eq:ce}
\end{equation}

For the recognition (i.e. identity classification) task, the state-of-the-art ArcFace loss was used. ArcFace showed impressive results when compared to the conventional softmax-loss due to the explicit enforcing of a higher intraclass similarity for and diversity for inter-class samples. Through the $l_2$ normalization of both the weights and the feature vector, this loss becomes equal to the geodesic distance margin penalty in a normalised hypersphere. The formulation for this loss can be seen in Equation~\ref{eq:arc_face}, where $m$ represents the margin, $s$ the scale, and $\theta_{it}$ the angle between features, $x_i$ and weights, $W_t$, for sample $i$ and target class $t$. Both $m$ and $s$ are hyperparameters of the loss.

\begin{equation}
    L_{arc} = -\frac{1}{N}\sum^{N}_{i=1}\log\frac{e^{s\left(\cos\left(\theta_{it} + m\right)\right)}}{e^{s\left(\cos\left(\theta_{it} + m\right)\right)} + \sum^{n}_{j=1, j \neq t}e^{s\cos\theta_{ij}}}
\label{eq:arc_face}
\end{equation}

\subsection{Contrastive learning}

The multi-task configuration of the network ensures that the network learns useful features to be aware of the mask presence. However, there are no guarantees that those features are leveraged by the main recognition branch. In fact, in practice, it can negatively impact the network performance whenever not carefully handled. Due to this, the proposed method introduces a contrastive training approach to promote mask awareness on the main branch. 

The contrastive training aims to give two images from the same individual to the same network. One of the images is augmented with synthetic masks, whereas the other has no mask augmentation. At loading time if one of the images is horizontally flipped, the other is also flipped. The produced face embeddings of both images are given to a loss function that calculates the distance between them. The formulation of this loss is seen in Equation~\ref{eq:mse}, where $F$ stands for the feature vector size, $xm_{ij}$ represents the value for masked feature vector $i$ at position $j$, whereas $xu_{ij}$ represents the same but for the unmasked feature vector. 

\begin{equation}
    L_{MSE} = \frac{1}{N}\frac{1}{F}\sum_{i=1}^{N}\sum_{j=1}^{F}{(xu_{ij} - xm_{ij})^2}
    \label{eq:mse}
\end{equation}

By enforcing smaller distances between masked and unmasked images, the network focus is promoted towards common features to both images. This also improves the capabilities of the model at the verification of masked references and masked probes. The selection of the augmented image was subject to careful evaluation. For this, the experiments consisted of training a model where the masked image was the masked version of the unmasked image and another where the masked image was a random masked version of any other image of the individual.

\subsection{Global architecture}

The overall architecture of the model comprises the previous two modules and a backbone convolutional neural network as seen in Figure~\ref{architecture_scheme}. For the backbone, the selected architectures were the 50 and 100 layers versions of the ResNet~\cite{he2016deep}. The last layer of these was replaced by two parallel fully-connected layers that produce two distinct embeddings. The first embedding with a size of 512 contains the facial features extracted for the verification task, whereas the second with a size of 32 is responsible for the mask detection. 

The modules introduced in this architecture can be used on top of already trained backbones. This leads to faster convergence times overall. To show that, we included in the experiments the pretrained ResNet-100 backbone called ElasticFace~\cite{boutros2021elasticface}.

To train the entire architecture end-to-end, it is necessary to combine the losses. First, the multi-task loss must be formulated, as seen in Equation~\ref{eq:branch}. The network runs twice, one with masked images and the other with unmasked images. Initially, this loss is computed independently for both runs. Afterwards, it is combined on the final loss. The hyperparameter $\lambda$ is responsible for controlling the impact that the mask detection has on the loss. 

\begin{equation}
    L_{masked/unmasked} = L_{arc} + \lambda L_{CE}
    \label{eq:branch}
\end{equation}

The masked and unmasked branches are combined on Equation~\ref{eq:combined}, which represents the formulation of the final loss. In this loss, the component regarding the $L_{MSE}$ has its impact controlled by the hyperparameter $\alpha$, whereas the impact of the combination of the two branches (masked and unmasked) is controlled by $\beta$. It is important to note that the weights are shared between the networks that are used to computed both masked and unmasked inputs, and that its optimization occurs simultaneously for all loss components.  

\begin{equation}
    L_{comb} = \alpha L_{MSE} * \beta (L_{masked} + L_{unmasked})
    \label{eq:combined}
\end{equation}

\section{Experimental Setup}
\label{sec:experimental_setup}
\subsection{Training details}

The backbone architectures to be trained for masked face recognition are the ResNet-100 and ResNet-50~\cite{he2016deep}. The selection of these architectures was motivated by the wide use of the first in the literature and state-of-the-art face recognition methods~\cite{deng2019arcface,an2021partial,huang2020curricularface,sun2020circle,duan2019uniformface}. Similarly to the literature~\cite{boutros2021elasticface}, on ResNet-100 we set the parameter $s$ to 64, and on the ResNet-50 we set it to 30. In both architectures, $m$ was set to 0.5. Due to GPU memory constraints, experiments that trained the backbone from scratch used a mini-batch size of 480. If the backbone was pretrained and frozen the mini-batch size increased to 512. The models in this paper are implemented using Pytorch~\cite{paszke2019pytorch} and ran on one Linux machine with 4 Nvidia Tesla V100 34 gb GPUs. All the experiments were trained with Stochastic Gradient Descent (SGD) with an initial learning rate of 1e-1, a momentum of 0.9 and weight decay of 5e-4. The learning rate is decreased by a factor of 10 at 34k and 64k iterations for models with pretrained backbones, and at 60k and 96k for the others. While loading the images at training, the image is horizontally flipped with a probability of 0.5. The networks expect 112x112x3 sized images and produce 512-d and 32-d embeddings. The latter is only computed at training time.  The images are pre-aligned and cropped using the Multi-task Cascaded Convolutional Networks (MTCNN)~\cite{zhang2016joint}. The images are also further normalized to have pixel values between -1 and 1. Regarding the hyperparameters of the loss, $\lambda$, $\alpha$ and $\beta$ are set to 0.1, 1/3 and 1/2, respectively.

\subsection{Training dataset}

Similarly to recent works in the literature~\cite{deng2019arcface,an2021partial,huang2020curricularface,meng2021magface}, we used the MS1MV2 dataset~\cite{deng2019arcface}. The MS1MV2 is a refined version of the original MS-Celeb-1M dataset~\cite{guo2016ms}. This refined dataset contains 5.8M images of 85k distinct identities. We augmented this dataset with face masks to twice its original size while not interfering with the identities. For these augmentations, we used the MaskTheFace open-source code~\cite{anwar2020masked}. The mask generation process is offline, thus not interfering with training times. MaskTheFace has five types of masks from which we used four (surgical, N95, KN95, cloth). The type of the mask and its colour were selected randomly from all the available options. A similar methodology was applied to all the probes in the Labeled Faces in the Wild (LFW) dataset~\cite{huang2008labeled}. Due to its wide use in the literature, the LFW dataset is used for the validation of the algorithm after each epoch.

\subsection{Evaluation benchmarks and metrics}

\begin{figure}[h!]
    \centering
  \begin{subfigure}[b]{0.4\linewidth}
       \includegraphics[width = \linewidth, height=2.7cm]{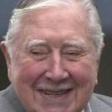}
       \caption{Example from the LFW dataset.}
       \label{unmaskedgood}
  \end{subfigure}
  \begin{subfigure}[b]{0.4\linewidth}
       \includegraphics[width = \linewidth, height=2.7cm]{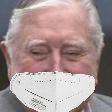}
       \caption{Example from the LFW dataset mask augmented.}
       \label{maskedgood}
  \end{subfigure}
  
  \begin{subfigure}[b]{0.4\linewidth}
       \includegraphics[width = \linewidth, height=2.7cm]{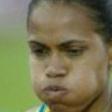}
       \caption{Example from the LFW dataset.}
       \label{unmaskedbad}
  \end{subfigure}
  \begin{subfigure}[b]{0.4\linewidth}
       \includegraphics[width = \linewidth, height=2.7cm]{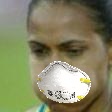}
       \caption{Example from the LFW dataset mask augmented.}
       \label{maskedbad}
  \end{subfigure}
  
  \caption{{Images extracted from the validation dataset for masked face recognition: images of the same individual with (a) and without a mask (b) with a accurate augmentation; images of the same individual with (c) and without a mask (d) with a poor augmentation.}}
  \label{fig:dataset} 
\end{figure}

We report the verification performance through the Equal Error Rate (EER), the area under the curve (AUC), FMR100 and FMR10. The latter two are the lowest FNMR for an FMR$\le$1.0\% and $\le$10.0\%, respectively. We further report the mean of genuine predictions (Gmean) and the mean of impostor predictions (IMean) to analyse the distribution of the scores and how distant are the means of these two categories. We evaluate our approach in two distinct settings. Firstly, the solution was evaluated in the U-M setting, which has unmasked references and masked probes. Afterwards, the evaluation in the M-M setting, where both probes and references are masked, was also conducted. The validation of the model and the model selection was carried on the LFW dataset.  Figure~\ref{fig:dataset} shows some examples of the images that are included in the dataset. The metric used to select the final model was the FMR100 on the U-M setting in the validation dataset. The receiver operating characteristic (ROC) curve was plotted for mask detection on the LFW dataset.

\begin{table*}
 \caption{Experiments on the distinct configurations of the proposed method on the test dataset (in the column ``Selection'': Original stands for the use of the masked version of the unmasked image on the second run of the network, whereas Random stands for the use of a randomly picked masked image from the same subject of the unmasked image;in the column ``Mode'': U and M stands for unmasked and masked data, respectively; EER, FMR100, FMR10 in $\%$).}
\label{res-rmfd}
\centering
\begin{tabular}{lccccccccc}
\toprule
Method    & Pretrained & Selection & Mode & GMean & IMean & AUC & EER & FMR100 & FMR10 \\ 
\midrule

 ResNet-50 & No & Original  & \makecell{U-M\\M-M} &\makecell{0.561\\0.625} & \makecell{0.333\\0.341}& \makecell{0.988\\0.983} & \makecell{5.377\%\\5.346\%} & \makecell{7.285\%\\6.546\%} &\makecell{4.519\%\\4.642\%} \\
 \midrule
 ResNet-100  & Yes & Original & \makecell{U-M\\M-M} &\makecell{0.571\\0.627} & \makecell{0.357\\0.365}& \makecell{0.987\\0.983} & \makecell{4.917\%\\5.005\%} & \makecell{5.986\%\\6.109\%} &\makecell{3.922\%\\3.882\%} \\
 \midrule
 ResNet-100 & Yes & Random & \makecell{U-M\\M-M} &\makecell{0.567\\0.625} & \makecell{0.354\\0.361}& \makecell{0.984\\0.984} & \makecell{5.371\%\\5.581\%} & \makecell{5.986\%\\6.184\%} &\makecell{4.991\%\\5.056\%} \\
 \midrule
 ResNet-100 & No & Original & \makecell{U-M\\M-M} &\makecell{0.624\\0.675} & \makecell{0.373\\0.383}& \makecell{\textbf{0.992}\\\textbf{0.992}} & \makecell{\textbf{4.594\%}\\\textbf{4.329\%}} & \makecell{\textbf{5.750\%}\\\textbf{5.509\%}} &\makecell{\textbf{2.582\%}\\\textbf{2.836\%}} \\
 \midrule
 ResNet-100  & No & Random & \makecell{U-M\\M-M} &\makecell{0.621\\0.671} & \makecell{0.371\\0.382}& \makecell{0.988\\0.991} & \makecell{5.164\%\\4.757\%} & \makecell{5.917\%\\5.695\%} &\makecell{4.197\%\\3.668\%} \\

\bottomrule
\end{tabular}
\end{table*}

\subsection{Test dataset}

While for sanity validation of the model performance the LFW dataset with synthetic masks was used, the values reported for the experiments are with respect to another database. The proposed solution, trained on synthetic data, is evaluated on a dataset containing real masks. This dataset was the one used in the masked face recognition competition~\cite{boutros2021mfr}. It was collected during 3 different days from 47 subjects. The images captured on the first day are used as references, whilst the probes were captured during the second and third sessions.

In each session, three videos were recorded using a webcam, while the subjects looked directly at the camera. Two of the videos included masks, whereas the third did not. An overlapping database, and the same capture and frame selection procedure is described  in~\cite{DamerBiosig2020,DamerIetbmt2021}. The references contain 470 unmasked images and 940 masked images. The probes contain 940 unmasked images and 1880 masked images. These distinct captures allowed for the evaluation of the U-M and M-M configurations.

\section{Results}
\label{sec:results}

In this section, we present the results achieved by our method and how it compares with the literature. Comparisons focus mostly on variations of the proposed method, on the method presented by Neto~\textit{et al.}~\cite{neto2021my} and the solutions submitted to the competition hosted by Boutros~\textit{et al.}~\cite{boutros2021mfr}.

\begin{table}[h!]
 \caption{Number of parameters of each module of the proposed approach. Includes the information regarding the use of each module for inference. Training parameters refers only to trainable parameters during training.  }
\label{module-parameters}
\centering
\begin{tabular}{l|c|c}
\toprule
Module  & \#Parameters & Inference  \\ 
\midrule
 ResNet-100 w/o Embedding Layer & 52,309,568 & Yes\\
 Face Recognition Embedding Layer&  12,846,080 & Yes \\
 
 Mask Detection Embedding Layer& 802,880 & No \\

 ArcFace Layer & 43,899,904 & No\\
 Mask Detection Fully-Connected Layer& 66 & No \\
 \midrule
 \midrule
 Total (Training w/o pretraining) & 109,858,498 & - \\
 Total (Training w/ pretraining) & 57,548,930 & - \\
 \midrule
 \midrule
 Total (Inference) & 65,155,648 & - \\
 
\bottomrule
\end{tabular}
\end{table}

We tested several variations of the proposed method. From using a smaller backbone network to adding the proposed modules on top of pretrained methods, and distinct image selection approaches to the contrastive learning module. Four of these variations are capable of beating the baseline method presented at the masked face recognition competition~\cite{boutros2021mfr} for the U-M setting. For M-M there are two less model capable of outperforming the baseline. Table~\ref{res-rmfd} shows the results of the several variations of the proposed method on the test dataset.  The best performance is achieved by a ResNet-100 backbone trained from scratch for masked face recognition with pairs of images where the masked image corresponds to the masked version of the unmasked image.  On the other hand, we show that with little training, it is possible to improve the performance of already existing models without changing the features learnt by them for conventional face recognition. In the majority of the solutions presented in the competition did use the ResNet-100, the training process required at least tuning 65 million parameters from the backbone and almost 44 million parameters from the ArcFace layer. By freezing the parameters of the backbone (except the last layer), we reduce the parameters that require training by roughly 47.6\%. All of this, while still beating the baseline results. When the backbone is frozen our approach trains $13,648,960 + 43,899,904 = 57,548,864$  parameters, from which  $43,899,904 + 802,880 = 44,702,784$ are removed on inference. Our approach parameters are shown in Table~\ref{module-parameters}, as well as their use on inference.

As expected, when selecting the masked image for the contrastive learning task, using the masked version of the unmasked image leads to slightly better results. However, it is worth noting, that while it performs worse than our best model, using random masked images from the same individual still leads to better results than the baseline for both M-M and U-M.

In order to compare to the competition solutions, we adopted a selection criterion based on the loss function used, the feature vector size, the input size and the dataset used. Using this criterion we selected the Top-3 solutions.  The solutions were named in the competition as A1\_Simple (Top-1), VIPLFACE-M (Top-2) and MaskedArcFace (Top-3). These solutions used augmented versions of the MS1MV2 dataset, trained on ArcFace loss, expect inputs size of 112x112x3 and return a feature vector of 512-d.

\begin{table}[h!]
 \caption{Comparison of the FMR100 of the methods presented in the MFR competition~\cite{boutros2021mfr}, a baseline and the proposed approach. Methods evaluated on the competition dataset for U-M face verification. (FMR100 in $\%$).}
\label{res-um-sota}
\centering
\begin{tabular}{lc}
\toprule
Method  & FMR100  \\ 
\midrule
 Neto~\textit{et al.}~\cite{neto2021my} & 28.252\% \\
 \midrule
 Baseline~\cite{boutros2021mfr}&  6.009\%   \\
 \midrule
 
 Top-1 solution~\cite{boutros2021mfr}& \textbf{5.538\%}  \\
 \midrule

 Top-2 solution~\cite{boutros2021mfr} & 5.681\% \\
 \midrule
 Top-3 solution~\cite{boutros2021mfr}& 5.687\% \\
 \midrule
 \midrule
 \textbf{Ours}& 5.750\%   \\
\bottomrule
\end{tabular}
\end{table}

In Table~\ref{res-um-sota} it is possible to see how the proposed method performs against the methods in the literature  and the baseline on U-M face verification. The proposed approach achieved better results than the baseline and Neto~\textit{et al.}~\cite{neto2021my} method. However, it had a minor disadvantage when compared to the competition solutions. Nonetheless, the performance of those solutions is close to the proposed method, and there are explanations for the lower results. The best performing solution, the only one with a performance improvement $\ge0.1$ percentage points, used a model that has a roughly 34\% larger number of parameters~\cite{boutros2021mfr}. The second-best solution used a distinct synthetic mask generation technique, which can heavily impact the behaviour of the final model. Finally, the third-best solution, accordingly to the description given in the MFR competition paper~\cite{boutros2021mfr}, seems to also apply synthetic masks to the test set. Allied with the non-deterministic behaviour, distinct random seeds that lead to different random initialization, and potentially distinct mask augmentations (due to randomness on mask generation) these factors can explain the performance difference between methods. Moreover, the second and third-best performing solutions have only minor improvements of $0.069$ and $0.063$ percentage points, respectively. 

\begin{table}[h!]
 \caption{Comparison of the FMR100 of the methods presented in the MFR competition~\cite{boutros2021mfr}, a baseline and the proposed approach. Methods evaluated on the competition dataset for M-M face verification. (FMR100 in $\%$).}
\label{res-mm-sota}
\centering
\begin{tabular}{lc}
\toprule
Method  & FMR100  \\ 
\midrule
 Neto~\textit{et al.}~\cite{neto2021my} & 23.507\% \\
 \midrule
 Baseline~\cite{boutros2021mfr}& 5.925 \%   \\
 \midrule
 
 Top-1 solution~\cite{boutros2021mfr}& 5.771\% \\
 \midrule

 Top-2 solution~\cite{boutros2021mfr} & 5.759\% \\
 \midrule
 Top-3 solution~\cite{boutros2021mfr}&  5.825\%\\
 \midrule
 \midrule
 \textbf{Ours}& \textbf{5.509}\% \\
\bottomrule
\end{tabular}
\end{table}

Differently from U-M, in Table~\ref{res-mm-sota} it is possible to see that the proposed method outperforms the other solutions in M-M. Not only that but it outperforms every solution by larger margins than the ones shown in the previous table. When compared with our method, 1\_Simple, VIPLFACE-M  and MaskedArcFace have performance drops of $0.262$, $0.250$ and $0.316$ percentage points, respectively. Which means that the proposed approach performs better by a margin at least $\ge0.250$ percentage points when compared to the remaning approaches. While this seems a secondary evaluation, it is possible to exploit the potential of our method on M-M by adding synthetic masks to the reference image.

\begin{figure}[!h]
\centering
\includegraphics[width=0.5\textwidth]{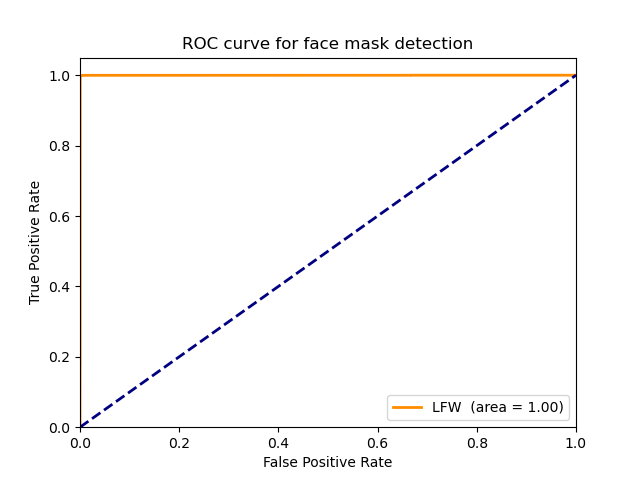}
\caption{Receiver operating characteristic (ROC) curve for mask detection on the LFW dataset with simulated masks.}
\label{roc_mask_detection}
\end{figure}

Due to the multi-task design of the architecture, it is possible to leverage the second embedding to do simultaneous mask detection. The mask detection performance of the proposed model was studied on the augmented LFW dataset in order to understand the feasibility of this task. We plotted the receiver operating characteristic (ROC) curve for this binary task (Fig.~\ref{roc_mask_detection}). Surprisingly the model achieves a perfect detection result on the validation dataset. The outstanding results on mask detection as well as the results of the masked face recognition task on both M-M and U-M show the effectiveness of the proposed method.

Propelled by the results shown on both tasks and by the effectiveness of the implementation of these modules without changing the weights of the backbone network (pretrained approach), we further argue in favour of the proposed method. Since the backbone is frozen during the training and only new embeddings are created it is possible, with just an increase of 13,648,960 parameters (roughly increasing the network size by  20\%), to compute simultaneously conventional face recognition, masked face recognition and mask detection embeddings. Thus, by leveraging the predictions of the masked detection embedding it is possible to decide between using the other two embeddings. Hence, avoiding any degradation to conventional face recognition, something that is not evaluated by the other methods in the competition. All of this, while still beating the baseline performance.

\section{Conclusion}
\label{sec:conclusion}
 
In this paper, we proposed and evaluated a novel multi-task approach based on contrastive learning to improve the verification accuracy on masked face recognition. The presented work was motivated by recent efforts on improving the verification accuracy of these models since there has been an increase in the use of face masks. The proposed solution is designed to work either on top of existing face recognition models or trained from scratch. It has been shown that both outperform the baseline. However, when running on top of existent methods, training and convergence are considerably lighter and faster.

The main goal of our FocusFace model is to promote the focus on areas that are not covered by face masks. To accomplish this, we include an additional embedding optimized to detect face masks. Hence optimizing the network to recognize face masks. To the multi-task module, we add a contrastive learning approach to enforce the network into constructing face embeddings that do not contain any mask related information. We explore the combination of three loss functions and two distinct input batches per iteration.

By experimenting with several variations of the architecture and comparing with distinct approaches, we demonstrated the effectiveness of our approach on improving the masked face verification performance. Moreover, we show that thanks to the capability of the proposed model to detect masks and to be implemented on top of current face recognition models, there is no need to sacrifice verification performance on conventional face recognition. Future work includes several variations to the contrastive learning module, a preprocessing step that adds synthetic masks to test data, and the study of larger and more accurate backbone systems.


{\small
\bibliographystyle{ieee}
\bibliography{egbib}
}

\end{document}